\documentclass[10pt,letterpaper]{article}
\usepackage[scheme=plain]{}

\usepackage{titlesec}
\usepackage{abstract}
\usepackage{cvpr}
\usepackage{caption}
\usepackage{times}
\usepackage{epsfig}
\usepackage{graphicx}
\usepackage{amsmath}
\usepackage{amssymb}
\usepackage{subfigure}
\usepackage{overpic}
\usepackage{enumitem}
\usepackage{booktabs}
\usepackage{subfigure}
\usepackage{fancybox}
\usepackage{fancyhdr}
\usepackage{lastpage}
\usepackage{longtable}
\usepackage{rotating}
\usepackage{multirow}
\usepackage{makecell}
\usepackage{float}
\usepackage[a4paper,top=3cm,bottom=2cm,left=3cm,right=3cm,marginparwidth=1.75cm]{geometry}
\usepackage[colorlinks=true, allcolors=black]{hyperref}
\usepackage{algorithm}
\usepackage{algorithmicx}
\usepackage{algpseudocode}

\cvprfinalcopy 
\title{Obtain Employee Turnover Rate and Optimal Reduction Strategy Based On Neural Network and Reinforcement Learning}
\author{Xiaohan Cheng \\
China Jiliang University\\
{\tt\small  lydiaecheng@outlook.com}}

\makeatletter
\renewcommand\section{\@startsection{section}{1}{\z@}%
{-3.5ex \@plus -1ex \@minus -.2ex}%
{2.3ex \@plus.2ex}%
{\textbf}}
\makeatother

\makeatletter
\renewcommand\subsection{\@startsection{subsection}{1}{\z@}%
{-3.5ex \@plus -1ex \@minus -.2ex}%
{2.3ex \@plus.2ex}%
{\textbf}}
\makeatother

\makeatletter
\renewcommand\subsubsection{\@startsection{subsubsection}{1}{\z@}%
{-3.5ex \@plus -1ex \@minus -.2ex}%
{2.3ex \@plus.2ex}%
{\textbf}}
\makeatother

\begin{document}

\maketitle

\textbf{Abstract:} Nowadays, human resource is an important part of various resources of enterprises. For enterprises, high-loyalty and high-quality talented persons are often the core competitiveness of enterprises. Therefore, it is of great practical significance to predict whether employees leave and reduce the turnover rate of employees. First, this paper established a multi-layer perceptron predictive model of employee turnover rate. A model based on Sarsa which is a kind of reinforcement learning algorithm is proposed to automatically generate a set of strategies to reduce the employee turnover rate. These strategies are a collection of strategies that can reduce the employee turnover rate the most and cost less from the perspective of the enterprise, and can be used as a reference plan for the enterprise to optimize the employee system. The experimental results show that the algorithm can indeed improve the efficiency and accuracy of the specific strategy.


\section{Introduction}

In today's society, employee loyalty is more and more important for the existence and development of a company. The high employee turn over rate not only may take away business and technical secrets, and takes away customers, the company suffers direct economic losses, but it also increases the cost of manpower replacement. It affects the continuity and quality of work, and also affects the stability of in-service employees. If it is not controlled, it will eventually affect the potential and competitiveness of the company's sustainable development. From the perspective of human resource management, the management of lost talents has a very important position in human resource management.

However, the personal data of employees is complicated and large in quantity, and the reasons for resignation are diversified. This poses a challenge for the human resources management department to predict whether employees will resign and make corresponding decisions. This paper aims to help companies find a series of strategies that are accurate, fast, and cost-effective to reduce the probability of employee turnover. Nowadays, most researches on employee turnover are still conducted from the aspects of behavioral psychology, management, etc., and rarely do mining and analysis directly from the level of employee data. However, mining and analysis from the data level often directly use methods such as logistic regression and random forest to directly classify employees according to the two states of resignation and non-resignation, which cannot describe the current status of employees more accurately and lost the dynamics.

This paper firstly defines the employee agent formally, and uses the multi-layer perception in the neural network to predict the employee turnover rate. The neural network can predict the turnover probability in real time based on the employee's feature set. Then based on the Sarsa reinforcement learning algorithm, the optimal solution strategy for this employee's resignation based on employee data is finally obtained. Reinforcement learning is inspired by the ability of organisms to effectively adapt to the environment, interacting with the environment through a trial and error mechanism, and learning the optimal strategy by maximizing accumulated rewards. Using this method can make up for the lack of data analysis in disciplines such as behavioral psychology and management.

\section{Method}

\subsection{Formal Definition of Employees}

The concept of Agent first appeared in Minsky's "Thinking Society" published in 1996. He believes that certain individuals in the society can obtain solutions to problems through negotiation, and individuals are Agents. The basic idea enables software to simulate human social behavior and cognition. So far, there is no unified Agent concept. In a nutshell, an agent is an entity that exists in a certain environment. It has the characteristics of autonomy, initiative, reactivity, intelligence, etc., and can perform reasoning, planning and other behaviors based on knowledge. 

Therefore, an employee agent can be formally defined in the following way.Employee Agent has characteristic attribute $C$, state attribute $S$, behavior set $A$, behavior cost set $COST$, and decision mechanism $D$, which are described as follows:

$$Agent = <C, S, A, COST, D>$$

The employee's characteristic attributes $C={c_0, c_1, c_2,...}$ can be possibly described as age, address, check-in times, etc. The employee's status attribute is $S \in [0, 1]$, where the larger $S$ indicates the greater the probability of resignation, and the smaller $S$ indicates the lower the probability of resignation. The employee behavior set $A \in \{a_0, a_1, ..\}$ defines a series of behaviors that may affect the change of S by the employee. These behaviors can be possibly described as recent salary increase, recent appreciation, recent business trip, etc. The employee's behavioral cost collection $COST={COST_1,COST_2, ...}$ corresponds to the employee's behavior collection $A$ one-to-one, describing the cost or price $COST_i$ paid by the company after the employee generates behavior $a_i$. The employee's decision-making mechanism $D$ mainly describes how the employee's state attribute $S$ and characteristic attribute $C$ will be changed after the employee has a behavior.

\subsection{Prediction of the Probability of Employee Turnover}

The data-type agent employee ignores other attributes of the employee agent, and only focuses on its feature set $C$. The employee feature set $C$ is digitized and standardized to form a data matrix. Table \ref{tab1} shows the unstandardized part of the data matrix in the sample data shared by the IBM Watson Analytics analysis platform.

\begin{table}[H]
	\centering
	\caption{Part of unstandardized data matrix.}
	\begin{tabular}{c|c|c|c}
		\toprule
        Employees/Features & Age & Attrition & DistanceFromHome \\
        \midrule
		$Employ_1$ & 37 & 0 & 1\\ 
        \hline
		$Employ_2$ & 54 & 0 & 1\\
        \hline
        $Employ_3$ & 34 & 1 & 7\\
        \bottomrule
	\end{tabular}
	\label{tab1}
\end{table}

\begin{figure}[H]
	\begin{center}
		\includegraphics[width=0.5\linewidth]{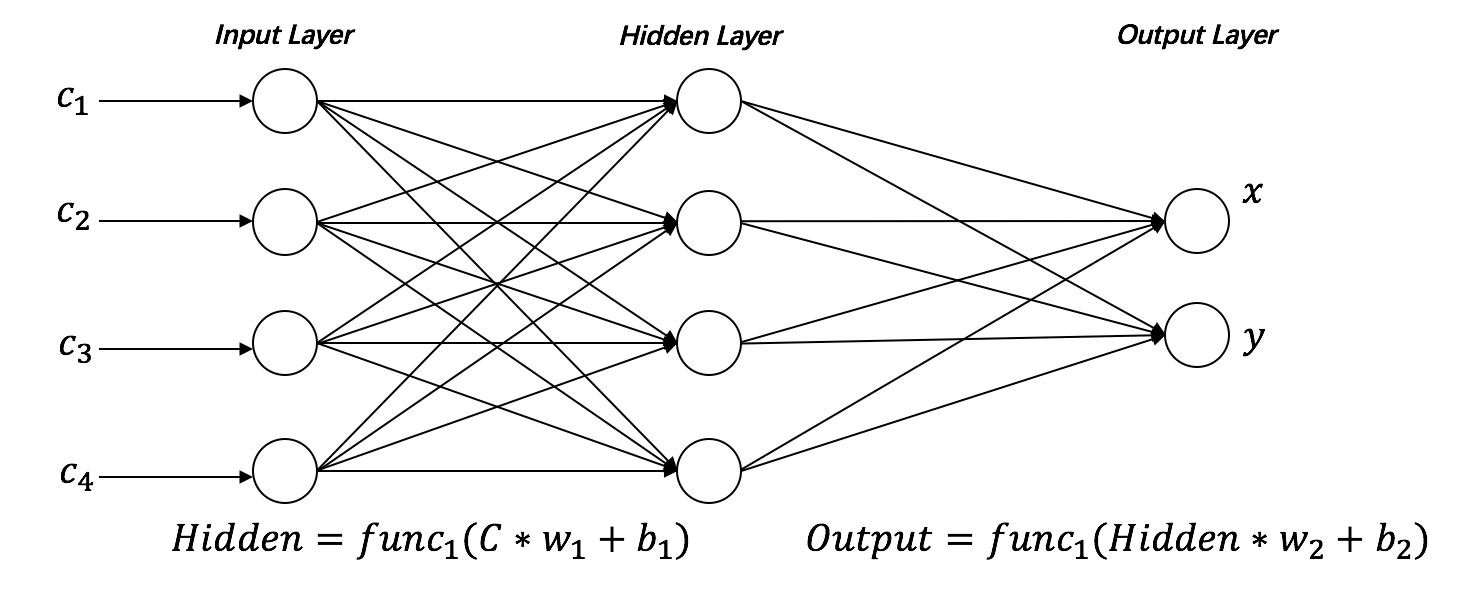}
		\caption{Multi-layer perception structure for binary classification problem.}
		\label{Fig:1}
	\end{center}
	\vspace{-0.5em}
\end{figure}

Multi-layer perception is one of the first neural networks proposed and the structure is shown in Figure \ref{Fig:1}. It is suitable for simple model classification tasks, such as two classification task. The employee turnover rate prediction problem is actually a two-category problem. The turnover label is $[1, 0]$ and the non-resignation label is $[0, 1]$. After inputting the information of the employee to be predicted, the predicted label $[x, y]$ can be directly calculated. We can simply regard the last two output neurons as a turnover factor and the other as a non-resignation factor. By comparing the relative value of the two, the probability of employee turnover can be obtained.

After obtaining the output layer $x$ and $y$, the employee turnover probability is $P_{out} = \frac{y}{x + y}$. In order to facilitate subsequent calculations, we use the non-resignation probability. Lowering the resignation probability means increasing the non-resignation probability. Probability of turnover: $P_{not\_out} = \frac{x}{x + y}$.

We saved the parameters of the neural network after the training, and combine the preprocessing function of the employee feature set $C$ (employee feature dataization, standardization), and encapsulate it into the $CalculateS$ function, and enter an employee feature set C, $C = {c_1, c_2, ...}$, output the employee's non-resignation probability $P_{not\_out} = CalculateS(C)$.

\subsection{Employee Turnover Rate Reduction Strategy based on Sarsa}
\subsubsection{Meta-Cost and Meta-Action}

This paper defines the concepts of meta-cost and meta-action, simplifies the process of reinforcement learning, and provides companies with more specific and easier-to-implement solutions. In order to reduce the probability of employee resignation, the company will make multiple decisions within a certain period of time, and multiple decisions will cause employees to produce multiple actions. The advantage is that the cost of each employee agent change is consistent, so we only need to make the employee’s non-resignation rate reach a certain threshold and take the shortest path. This set of strategies will take lowest cost.

First, suppose that there is a type of meta-action in the action set, and the cost of the action is at a moderate level in the action set. For example, the meta-action can be to promote a first-level position, go out for a one-month study, etc., rather than increase the employee's salary by one dollar, or let the employee go abroad for exchanges ten times.

The cost that needs to be consumed to carry out a meta-action, we define it as the meta-cost. Meta-cost and meta-action aim to standardize the description of employee's actions, so that each action in the action set can be transformed into meta-action according to the corresponding relationship. For example, if it costs 2 dollars for a trip abroad, we can redefine the action of studying abroad once as studying abroad 0.5 times, etc. After the standardized description, the cost consumed by each action is the same, but the effect of reducing the turnover rate on employees is different. On this basis, the shortest path from the initial turnover rate to the final turnover rate is the cost. Minimal solution strategy.

Therefore, based on the meta-action and meta-cost, we reformatically describe the employee agent: employee agent has characteristic attribute $C$, state attribute $S$, meta-cost $pCost$, meta-action set $A$, decision mechanism $D$, which means $Agent = <C, S, pCost, A, D>$

Among them, decision-making mechanism $D$ mainly acts on $S$ based on meta-action $A$, so it can be implicit in $A$, so it can be ignored and re-described as $Agent = <C, S, pCost, A>$

\subsubsection{Task Definition}

For an $Agent_i = <C, S, pCost, A>$, the Agent has a starting state $S_{start}$ and an expected final state $S_{end}$, and we want a shortest set of meta-action sequences $Ans$, where $Ans=\{ans_1, ans_2, ..., ans_n\}, ans_i \in A$, then $Ans$ is the final output sequence, and $pCost*|Ans|$ is the minimum total cost.

The existing method $S_{next} = CalculateS(Agent<C, S, A>[C])$, enter a feature set $C$ after the agent has taken action $A$, and return the agent's next state $S_{next}$.

\subsubsection{Sarsa Algorithm and Improvement}

Sarsa algorithm is an algorithm for learning Markov decision process strategies, used in the reinforcement learning field of machine learning. It was proposed by Rummery and Niranjan in 1994, and the name is "Modified Connectionist Q-Learning" (MCQ-L) \cite{Rummery1994On}. The name Sarsa reflects the fact that the main function of updating the Q value depends on the current state of the agent $S_1$, the agent chooses the action of $A_1$, and the agent gets the reward $R$ for choosing the action. The state $R_2$ entered after taking this action, and the next action $A_2$ selected by the Agent in its new state. The acronym for $(st ,at, rt, st+1, at+1)$ is sarsa. In the Sarsa algorithm, the strategy followed when selecting an action is the same as the strategy followed when updating the action value function, that is, the $ \epsilon $ -greedy strategy. This method is called On-Policy\cite{BradfordReinforcement}. The pseudo code is as follows.

\begin{algorithm}[H]
	\caption{Sarsa algorithm}
    \begin{algorithmic}[1]
    	\Require $Q(s, a),  \forall s \in S, a \in A, Q(terminal-state, ·) = 0$
 		\Repeat (for each episode)
        	\State Initialize $S$
            \State Choose $A$ from $S$ using policy derived from $Q$
            \Repeat (for each step of episode)
            	\State Take action $A$, observe $R$, $S'$
                \State Choose $A'$ from $S'$ using policy derived from $Q$
               	\State $Q(S,A) \gets Q(S,A) + \alpha [R + \gamma Q(S',A') - Q(S,A)]$
                \State $S \gets S'; A \gets A';$
            \Until S is terminal
        \Until end episode
    \end{algorithmic}
\end{algorithm}

The Sarsa algorithm mentioned above is for the general situation. Under the environment of this question, some improvements are needed. The state set of the employees in this question is the employee turnover rate. Employees need to find the shortest action from the current turnover rate to the ideal turnover rate in the entire environment. set. So every time an action is taken, the reward given and the state transition need to be improved.

Due to the characteristics of neural networks, the effects of meta-actions taken by employees in different states are not the same, and employees naturally transfer to the new state of neural network output after meta-actions are taken. The same meta-action may greatly reduce the employee turnover rate when employees are in different states, or may not reduce much. The cost of each meta-action is the same, so it is hoped that after each meta-action taken by the employee, the turnover rate can be reduced even more. Therefore, the reward can be directly described as the difference between the previous and the previous state. In each of its future training, in order to obtain more rewards, it has to take more meta-actions that can reduce the turnover rate, and then after training enough times, in every state of the employee, it can find The greater the meta-action that reduces the employee turnover rate, the faster the transition from the starting state to the ideal state. Therefore, the improved employee status transfer and reward can be defined as shown in the following formula.

$$S' \gets CalculateS(<C, S, A>)$$

$$R \gets (S'-S) $$

At the same time, the improved Sarsa algorithm can be obtained. Compared with the original Sarsa algorithm, we have introduced the environment discussed in this topic. In each training, we will take a meta-action that makes the difference between the front and back states larger. After the training reaches sufficient accuracy, we can find a shorter one. The set of actions as output.

\begin{algorithm}[H]
	\caption{Improved Sarsa algorithm}
    \begin{algorithmic}[1]
    	\Require $Q(s, a),  \forall s \in S, a \in A, Q(terminal-state, ·) = 0$
 		\Repeat (for each episode)
        	\State Initialize $S$
            \State Choose $A$ from $S$ using policy derived from $Q$
            \Repeat (for each step of episode)
            	\State Take action A
                \State $S' \gets CalculateS(<C,S,A>)$
                \State $R \gets (S'-S) $
                \State Choose $A'$ from $S'$ using policy derived from Q
               	\State $Q(S,A) \gets Q(S,A) + \alpha [R + \gamma Q(S',A') - Q(S,A)]$
                \State $S \gets S'; A \gets A';$
            \Until $S$ is terminal
        \Until end episode
    \end{algorithmic}
\end{algorithm}

\subsubsection{Strategy Generation}

There is a total of $Agent<C_{all}, S_{all}, A_{all}, pCost_{all}>$, select the i-th Agent among them, $Agent_i<C, S, A, pCost>, C \in C_{all}, S \in S_{all}, A \in A_{all}, pCost \in pCost_{all}$, existing methods $CalculateS(Agent<C, S, A>) = S_{next} $.

Step1: Initialize the Q-table $Q_{ij} = 0, i=1,2,..,|S_{all}|, j = 1,2,..,|A_{all}|$

Step2: Start training and train t times.

Step3: Select the random meta-action $a=random(A_{all})$ for the first time, and set the intermediate state $S_{temp} = CalculateS(Agent<C, S_{start}, a>)$, but select the meta-action $a$ corresponding to $A_{all}[argmax(Q_{S_{temp}})]$ when it is not the first time.

Step4: Calculate $S_{next} = CalculateS(S_{temp})$, $R = S_{next} - S_{temp}$

Step5: Calculate the predicted value $Q_{predict} = Q(S_{temp}, a)$

Step6: Twice select the meta-action $a_2 = A_{all}[argmax(Q_{S_next})]$

Step7: Calculate the target value $Q_{target} = R + \gamma Q(S_{next}, a_2)$

Step8: Update Q-table

$$Q(S_{temp}, a) = Q(S_{temp}, a) + \alpha [Q_{target} - Q_{predict}]$$

Step9: Update intermediate state $S_{temp} =  S_{next}$, if $S_{temp}$ equals $S_{end}$ then enter Step10, if not, then return to Step3 to continue.

Step10: Whether the number of training times has reached $t$ times, if yes, enter Step11, if not, set $S_{temp}=S_{start}$ and return to Step3 to train again.

Step11: Initialize the final result empty set $Ans$, execute Step3 again, add $a$ in each cycle to $Ans$, output $Ans$, and end.

\section{Experiment}

\begin{figure}[H]
	\begin{center}
		\includegraphics[width=0.5\linewidth]{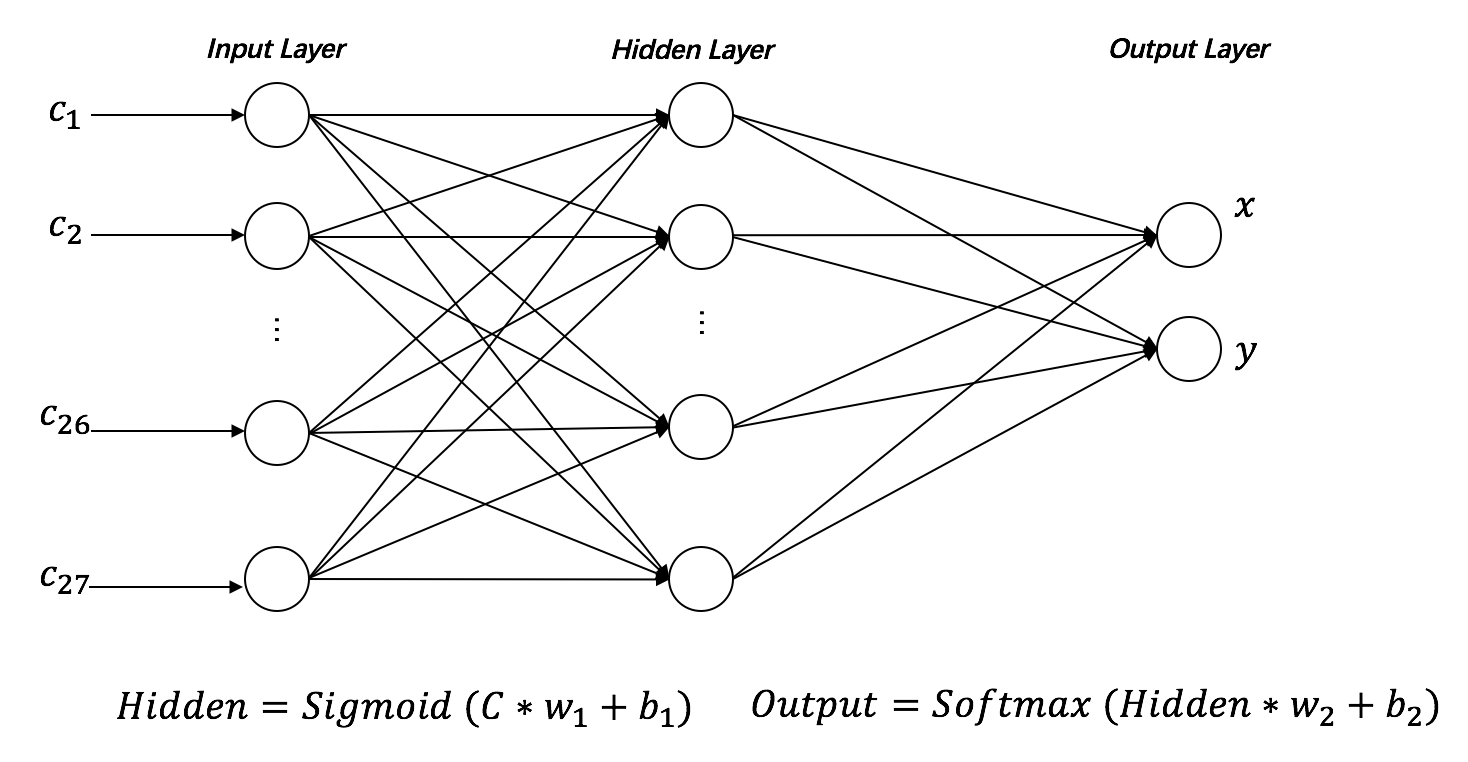}
		\caption{Our multilayer perception structure.}
		\label{Fig:2}
	\end{center}
	\vspace{-0.5em}
\end{figure}

In order to verify the feasibility and effect of the algorithm, this paper directly uses the sample data shared by the IBM Watson Analytics analysis platform for processing. The sample data contains 1101 employees, 80\% of which are selected as the training set and 20\% as the test set. Each employee contains 27 characteristics and 1 label. Construct a multi-layer perception model with 27 nodes in the input layer, 20 nodes in the hidden layer, and 2 nodes in the output layer. Let the input matrix be $X$, define the parameters $\{w_1, b_1, w_2, b_2\}$ to construct a multilayer perception as shown in Figure \ref{Fig:2}. The results are shown in \ref{tab2}.

\begin{table}[H]
	\centering
	\caption{Accuracy of employee turnover prediction obtained by multi-layer perception.}
	\begin{tabular}{c|c}
    	\hline
        Accuracy on training set & 96.41\% \\
		\hline
        Accuracy on testing set & 94.96\% \\
        \hline
	\end{tabular}
    \label{tab2}
\end{table}

Then, we randomly select employees $E<C,S,pCost,A>$, the current feature set part is shown in the Table \ref{tab3}. The initial non-resignation rate of this employee is 0.65, and we expect the non-resignation rate of this employee to eventually reach 0.80. Define the meta cost as 500 dollarss, and simply define the meta actions of the employee based on the meta-cost as shown in the Table \ref{tab4}. The cost of each action is 500 dollars.

\begin{table}[H]
	\centering
	\caption{Some characteristic attributes of the employee.}
	\begin{tabular}{c|c}
		\toprule
        Healthy value & 46 \\
        \hline
		Distance from home to company & 21 \\ 
        \hline
		Education level & 2\\
        \hline
        Environmental recognition & 4\\
        \hline
        Position & 2 \\
        \hline
        Job satisfaction & 2\\
        \hline
        Percentage increase in wages & 22\\
        \hline
        Interpersonal satisfaction & 4\\
        \bottomrule
	\end{tabular}
    \label{tab3}
\end{table}

\begin{table}[H]
	\centering
	\caption{The employee's meta-action collection.}
	\begin{tabular}{c|c|c}
    	\toprule
        meta-actions & Employee characteristics & Value after each action \\
		\midrule
        Get more exercise & Healthy value & 0.5 \\
        \hline
		Add some shuttles & Distance from home to company & 0.1 \\ 
        \hline
		Go out for a some training courses & Education level & 0.05\\
        \hline
        Improve environmental recognition & Environmental recognition & 0.05\\
        \hline
        Promotion & Position & 0.05 \\
        \hline
        Improve professional satisfaction & Job satisfaction & 0.05\\
        \hline
        Raise salary & Percentage increase in wages & 1.5\\
        \hline
        Strengthen interpersonal cooperation & Interpersonal satisfaction & 0.5\\
        \bottomrule
	\end{tabular}
    \label{tab4}
\end{table}

Input the improved Sarsa algorithm to train one hundred times, and the comparison between the first training result and the hundredth training result can be obtained. Thirteen strategies were given for the first training with a total cost of 6,500 dollars, and seven strategies were given for the 30th training with a total cost of 3500 dollars. Compared with other related works that study turnover strategies, which mainly focus on the composition of turnover costs, a large number of strategies for compensation and management are given in different stages. As a result, a clearer solution is given, which is to use neural network to comprehensively consider the employee's data to predict the turnover rate, and then the employee's turnover rate is given, and the best fixed cost strategy is given. The advantages of clarity and better cost control.

In the same way, it can be compared with the research results of the existing data mining technology on human resource management, the existing general strategy is to use decision trees, etc. The clustering algorithm clusters the position levels of employees, and prescribes the right medicine according to different categories and the basis of judgment during clustering. The comparison is shown as follows.

\begin{table}[H]
	\centering
	\caption{Strategies derived from cluster analysis.}\label{tab5}
    \begin{tabular}{c|c} 
    \toprule
    Targets & Employee collection after clustering \\
    \midrule 
    Type 1 employees & Organize more collective activities and care for employees \\
     \hline
    Type 2 employees & Give full play to talents and stimulate potential \\
     \hline
    Type 3 employees & Improve work motivation\\
     \hline
    Type 4 employees & Improve communication and organization skills \\
     \bottomrule
	\end{tabular}
\end{table}

\begin{table}[H]
	\centering
	\caption{Neural network combined with reinforcement learning strategies.}\label{tab5}
    \begin{tabular}{c|c}        
       \toprule
    Targets & Single Employee\\
    \midrule
    Non-turnover rate 66\% & Improve career satisfaction \\
     \hline
    Non-turnover rate 72\% & Improve interpersonal cooperation \\
     \hline
    Non-turnover rate 72\% & Go out for training courses \\
    \hline
    Non-turnover rate 79\% & Improve interpersonal cooperation \\
     \bottomrule
	  \end{tabular}
      \caption{Neural network combined with reinforcement learning strategies}
\end{table}

\begin{table}[H]
	\centering
	\caption{Strategy comparison}
	\begin{tabular}{c|c}
		\toprule
        Cluster analysis &  Neural network combined with reinforcement learning  \\
        \midrule
		Cannot target a single specific employee & Can target a single specific employee \\ 
        \hline
		Unable to target specific corporate environment & Can be targeted to specific corporate environments\\
        \hline
        Inability to control costs & be able to control costs \\
        \hline
        Unable to automatically generate strategy & Strategy can be generated automatically \\
        \bottomrule
	\end{tabular}
    \label{tab5}
\end{table}

By comparison, the method described in this paper is more targeted, directly targeting a single employee for dynamic strategy generation, so it has greater flexibility. Compared with the traditional cluster analysis, the method in this paper has stronger fit and clearer data and standards, so it is more flexible and more convenient for business managers to control. At the same time, due to the potential strong clustering relationship in the company, the method provided in this paper does not consider it. In the future follow-up development, you can first perform cluster analysis on employees from a higher level, and then use neural network joint reinforcement learning methods The accuracy of generating strategies may continue to improve.

\section{Conclusion}

This paper proposes a strategy for reducing employee turnover rate based on reinforcement learning, which aims to help company seniors or human resource managers find a set of strategies with the lowest cost and effectiveness to retain employees. Firstly, the employees are described in the Agent form, the employee turnover rate and non-resignation rate are calculated by training the multilayer perception neural network, and the trained neural network is saved. And the concept of meta cost and meta action is proposed, and the Sarsa reinforcement learning algorithm is improved to find the shortest meta action set from the current non-resignation rate to the ideal non-resignation rate, which is the strategy set. Finally, a demonstration was carried out using the data set shared by the IBM Watson Analytics analysis platform. The employee turnover rate can be obtained simply and efficiently, and a series of strategies to reduce the cost of reducing the employee turnover rate are given.

\clearpage
\bibliographystyle{ieeetr}
\bibliography{main}

\end{document}